\documentclass{article}
\usepackage{graphicx}
\usepackage{subcaption}

\usepackage{hyperref}
\usepackage{amsmath}
\usepackage{amsfonts}
\usepackage{amsthm}
\usepackage{amssymb}

\usepackage[accepted]{icml2018}

\icmltitlerunning{Random Hinge Forest for Differentiable Learning}

\newcommand{\bx}{\mathbf{x}}
\newcommand{\bt}{\mathbf{t}}
\newcommand{\bF}{\mathbf{f}}

\newcommand{\bw}{\mathbf{w}}
\newcommand{\bz}{\mathbf{z}}
\newcommand{\bzero}{\mathbf{0}}
\newcommand{\reals}{\mathbb{R}}

\newcommand{\relu}{\text{ReLU}}
\newcommand{\leaf}{\ell}
\newcommand{\loss}{\mathcal{L}}
\newcommand{\sgn}{\text{sgn}}

\newcommand{\argmin}{\operatornamewithlimits{argmin}}
\newcommand{\ie}{\textit{i}.\textit{e}.}
\newcommand{\eg}{\textit{e}.\textit{g}.}

\begin{document}
\twocolumn[
\icmltitle{Random Hinge Forest for Differentiable Learning}
\icmlsetsymbol{equal}{*}

\begin{icmlauthorlist}
  \icmlauthor{Nathan Lay}{}
  \icmlauthor{Adam P. Harrison}{}
  \icmlauthor{Sharon Schreiber}{}
  \icmlauthor{Gitesh Dawer}{fsu}
  \icmlauthor{Adrian Barbu}{fsu}
\end{icmlauthorlist}

\icmlaffiliation{fsu}{Department of Statistics, Florida State University, Tallahassee, FL}
\icmlcorrespondingauthor{Nathan Lay}{enslay@gmail.com}

\icmlkeywords{Random Forest, Random Fern, Differentiable Learning, Computation Graphs}

\vskip 0.3in
]

\printAffiliationsAndNotice{}

\begin{abstract}
We propose random hinge forests, a simple, efficient, and novel variant of decision forests. Importantly, random hinge forests can be readily incorporated as a general component within arbitrary computation graphs that are optimized end-to-end with stochastic gradient descent or variants thereof. We derive random hinge forest and ferns, focusing on their sparse and efficient nature, their min-max margin property, strategies to initialize them for arbitrary network architectures, and the class of optimizers most suitable for optimizing random hinge forest. The performance and versatility of random hinge forests are demonstrated by experiments incorporating a variety of of small and large UCI machine learning data sets and also ones involving the MNIST, Letter, and USPS image datasets. We compare random hinge forests with random forests and the more recent backpropagating neural decision forests.
\end{abstract}

\section{Introduction}
Random forest~\cite{breiman2001random} is a popular and widely used ensemble learning algorithm. With constituent models being decision trees, its formulation offers some level of interpretability and intuition. It also tends to generalize well with modest parameter tuning for a variety learning tasks and data sets. However, the recent success of deep artificial neural networks have revealed limitations of random forest and similar greedy learning algorithms. Deep artificial neural networks have exhibited state-of-the-art and even super human performance on some types of tasks. The success of deep neural networks can at least be partly attributed to both its ability to learn parameters in an end-to-end fashion with backpropagation and its ability to scale on very large data sets. By contrast, random forest is limited to fixed features and can plateau on test performance regardless of the training set size. Deep neural networks, however, are difficult to develop, train and even interpret. As a consequence, many researchers shy away from exploring and developing their own network architectures and often use and tweak pre-trained off-the-shelf models (for example~\cite{simonyan2014very,ronneberger2015u,he2016deep}) for their purposes.

We propose a novel formulation of random forest, the random hinge forest, that addresses its predecessor's greedy learning limitations. Random hinge forest is a differentiable learning machine for use in arbitrary computation graphs. This enables it to learn in an end-to-end fashion, benefit from learnable feature representations, as well as operate in concert with other computation graph mechanisms. Random hinge forest also addresses some limitations present in other formulations of differentiable decision trees and forests, namely efficiency and numerical stability. We show that random hinge forest evaluation and gradient evaluation are logarithmic in evaluation complexity, compared to exponential complexity like previously described methods, and that random hinge forest is robust to activation/decision function saturation and loss of precision. Lastly, we show that random hinge forest is better than random forest and comparable to the state-of-the-art decision forest, neural decision forest~\cite{kontschieder2015deep}.

This paper first describes a series of related works and how they are different, then formulates the random hinge tree, ferns and forest. Then a series of experiments and results are presented on UCI data sets and MNIST. We compare the performance of the proposed methods with random forest~\cite{breiman2001random} and neural decision forest~\cite{kontschieder2015deep}, discuss the findings and finally conclude the paper.

\section{Related Work}
The forerunner of this work is random forest which was first described by~\cite{amit1997shape} and later by~\cite{breiman2001random}. Random forest is an ensemble of random decision trees that are generally aggregated by voting or averaging. Each random tree is trained on a bootstrap-aggregated training sample in a greedy and randomized divide-and-conquer fashion. Random trees learn by optimizing a task-specific \textit{gain} function as it divides and partitions the training data. Gain functions have been defined for both classification and regression as well as a plethora of specialized tasks such as object localization~\cite{gall2013class,criminisi2010regression}. Trees are also found as components used in other learning algorithms such as boosting~\cite{friedman2001greedy}. 

The works of~\cite{suarez1999globally,kontschieder2015deep}  make partitioning thresholds \textit{fuzzy} by representing the threshold operation as a differentiable sigmoid. The objective is to make a decision tree that is differentiable for end-to-end optimization. When traversing these fuzzy trees, each decision will result in a degree of \textit{membership} in the left and right partitioned sets. For example, if the decision produces a value of $0.4$, then the decision results in $0.4$ membership in one partition and $0.6$ membership in the other. When the decision process is taken as a whole, each resulting partition will have a membership that is a generally defined as a path-specific product of these decisions producing, however small, non-zero membership for all partitions. The work of~\cite{kontschieder2015deep} also introduces an alternating step for training the predictions and decisions and goes one step further and describes simultaneously optimizing all trees of a forest in an end-to-end fashion. The work~\cite{schulter2013alternating} globally optimizes a forest against an arbitrary loss function in a gradient boost framework by gradually building the constituent trees up from root to leaf in a breadth-first order.

The work of~\cite{ozuysal2010fast} develops a constrained version of a decision tree known as a random fern. A single random fern behaves like a checklist of tests on features that each produce a yes/no answer. Then the combination of yes/no answers is used to look up a prediction. Unlike random forest that are aggregated by averaging or voting, random ferns are aggregated via a semi-na{\"i}ve Bayes formulation. And unlike greedy learning decision trees, each random fern is trained by picking a random subset of binary features and then binning the training examples in the leaves. These leaves are then normalized to be used in a semi-na{\"i}ve Bayes aggregation.

Random hinge forest also resembles the models learned by multivariate adaptive regression splines (MARS)~\cite{friedman1991multivariate}. MARS is a learning method that fits a linear combination of hinge functions in a greedy fashion. This is done in two passes where hinge functions are progressively added to fit the training data in the first pass, and then some hinge functions are removed in the second pass to help produce a generalized model. 

Lastly, the work of~\cite{krizhevsky2012imagenet} and an abundance of works that follow, all very successfully used rectified linear units (ReLU) as activation functions. This was done for training efficiency and to keep non-linearity. It also, to some extent, helps prevent saturation problems that lead to near-zero gradients with sigmoid activations. To the best knowledge of the author, differentiable decision trees have always relied on sigmoid-like decision functions that resemble the activation functions of artificial neural networks. This reliance of sigmoid decision functions in fuzzy decision trees, the plethora of numerical problems with sigmoid activation functions, and the very successful use of ReLU in~\cite{krizhevsky2012imagenet} provided clues to formulate random hinge forest.

Random hinge forest is different from random forest in that constituent trees are all inferred simultaneously in an end-to-end fashion instead of with greedy optimization per tree. However, the trees of random forest can currently learn decision structures not easily described by random hinge trees as formulated here. And where random trees deliberately choose splitting features that optimize information gain, random hinge trees indirectly adjust learnable splitting features to fit their randomized decision structure. Random hinge forest is also different from the works of~\cite{suarez1999globally,kontschieder2015deep} and fuzzy decision trees in general in that random hinge trees use ReLUs for decision functions instead of sigmoids. This retains the desirable evaluation behavior of decision trees. Where fuzzy decision trees admit a degree of \textit{membership} to several partitions of the feature space, random hinge trees admit \textit{membership} to only one partition as with \textit{crisp} decision trees. This also implies that for a depth $D$ tree, random hinge trees need only evaluate $D$ decisions, while fuzzy trees need to evaluate on the order of $2^D$ decisions. Like~\cite{kontschieder2015deep}, random hinge forest can simultaneously learn all trees in an end-to-end fashion. However, random hinge forest is not limited to probability prediction or leaf purity loss and does not need alternating learning steps for leaf weights and thresholds and can be trained in the usual forward-backward pass done in computation graphs. The work of~\cite{schulter2013alternating} also builds up trees in an iterative fashion while random hinge forest assume complete trees and a randomized initial state. Both random hinge forest and alternating decision forest aim to optimize global loss functions, but the former can operate as part of a general computation graph. This work also presents the random hinge fern which bares the same decision constraint as random fern, but we have not explored semi-na{\"i}ve Bayes aggregation. This work also provides a way to train the random hinge fern in an end-to-end fashion which, to the best of our knowledge, has never been done for random ferns. 

\vspace{-1mm}
\section{Random Hinge Trees and Ferns}
Decision trees represent piecewise constant functions where each piece is disjoint from the rest and represented as a leaf of a decision tree. Each leaf is defined as a conjunction of conditions. Suppose each leaf $\leaf \subseteq T$ of vertices $v \in T$ in tree graph $T$ represents a unique path to leaf $\leaf$, then the conjunction to a given leaf $\leaf$ is defined as an indicator function
\vspace{-2.0mm}
\begin{equation}
I_{\leaf}(\bx) = \bigwedge_{(d,v) \in \leaf} I(d(\bx_{f_v} > t_v))
\label{eq:treeConj}
\vspace{-2.0mm}
\end{equation}
Where $d(x)$ represents the direction of traversal where
\vspace{-2.0mm}
\begin{equation}
d(x) = \begin{cases}
\neg x & \text{the next vertex is the left child} \\
x & \text{otherwise}
\end{cases} 
\vspace{-2.0mm}
\end{equation}
and $f_v, t_v$ are the feature index and threshold for the decision at vertex $v$. And since all conjunctions are disjoint, then a decision tree prediction can be represented as a summation over its leaves
\vspace{-2.0mm}
\begin{equation}
h_T(\bx) = \sum_{\leaf \in T} w_{\leaf} I_{\leaf}(\bx) 
\label{eq:treeSum}
\vspace{-2.0mm}
\end{equation}
We additionally abuse notation for $\leaf$ and treat it as an integer index where $w_{\leaf}$ is an arbitrary task-specific prediction (e.g. a tensor) for leaf $\leaf$ stored in component $\leaf$ of $w$. Since $I(\bx_{f_v} > t_v)$ has a derivative of $0$ almost everywhere, this form is not useful for optimization with gradient-based methods. Past works~\cite{suarez1999globally,kontschieder2015deep} that optimize decision trees with gradient descent approximated $I(\cdot)$ with a sigmoid function so that it would have non-zero derivatives. Sigmoid functions are never zero and thus all the terms of~\eqref{eq:treeSum} are non-zero when using sigmoids. Efficient implementation of decision trees have an evaluation operation that requires on the order of depth of the tree operations, but such \textit{fuzzy} approximations may require on the order of the total number of vertices to evaluate. 

Random hinge trees draw from the curious Rectified Linear Unit (ReLU) usage of~\cite{krizhevsky2012imagenet} as inspiration. However, ReLU cannot be used in a synonymous manner as the sigmoids from fuzzy decision trees. ReLU is not bounded in $[0,1]$ which is generally used in fuzzy logic. Instead, we imagine a pretend logic where $\text{True} = \{ x > 0 : x \in \reals \}, \text{False} = \{ x < 0 : x \in \reals \}$. The special value $x = 0$ is ambiguous and is neither part of the True nor False set. From there, we can define consistent logic operations like $\wedge, \vee, \neg$ as
\vspace{-2.0mm}
\begin{align}
A \wedge B &= \min \{A,B\} \\
A \vee B &= \max \{A,B\} \\
\neg A &= -A 
\end{align} \\[-6.0mm]
Then we define the binary relation $A>B$ as
\vspace{-2.0mm}
\begin{equation}
A>B = \max \{A-B, 0\} = \relu(A-B)
\vspace{-2.0mm}
\end{equation}
We could choose other definitions for $A>B$ like, for example, $A>B = A-B$, however this definition would suffer the same efficiency problem as fuzzy decision trees since it is non-zero even when it expresses values in the False set. Additionally, its linear nature may limit the tasks it can solve. By \textit{casually} substituting this \textit{logic} and binary relation into~\eqref{eq:treeConj}~\eqref{eq:treeSum}, we get
\vspace{-2.0mm}
\begin{align}
\hat{I}_{\leaf}(\bx) &= \min_{(d,v) \in \leaf} \{ \relu(d(\bx_{f_v} - t_v)) \} \label{eq:treeConjHinge} \\
h_T(\bx) &= \sum_{\leaf \in T} w_{\leaf} \hat{I}_{\leaf}(\bx) \\
&= \sum_{\leaf \in T} w_{\leaf} \min_{(d,v) \in \leaf} \{ \relu(d(\bx_{f_v} - t_v)) \} \label{eq:treeSumHinge} 
\end{align} \\[-4.0mm]
Here $\hat{I}_{\ell}(\cdot)$ denotes this \textit{altered} indicator function. We note that~\eqref{eq:treeConjHinge} does not actually use the logical complement $\neg (A>B) = -\text{ReLU}(A-B) \in \{0\} \cup \text{False}$, but instead uses $A<B = \text{ReLU}(B-A) \in \{0\} \cup \text{True}$. The summation over leaves in~\eqref{eq:treeSumHinge} is differentiable almost everywhere and preserves the efficiency of conventional decision trees in that at most a single term of~\eqref{eq:treeSumHinge} is ever non-zero. And as this form does not involve any product of sigmoid functions it does not suffer saturation issues nor does it suffer any potential loss of precision from multiplying many values in $(0,1)$. The traversal algorithm used for decision tree prediction can be used with some modification where the decision margin is also tracked during traversal. Such a traversal algorithm is described in algorithm~\ref{alg:computeKeyAndSignedMarginTree} and would result in~\eqref{eq:treeSumHinge} reducing to $h_T(\bx) = w_{\leaf} |r^*|$, where $\leaf,\ r^*$ are the result of the traversal algorithm.

In the context of classification, random hinge tree can be interpreted as a min-max margin classifier. The smallest margin along the path to a leaf is a factor of the output prediction. When this margin is near zero, the tree produces a corresponding prediction near zero indicating uncertainty. This reveals this type of tree to be extremely \textit{pessemistic}. For regression tasks, a random hinge tree is a piecewise linear model comprised of hinge functions.

The derivatives of $h_T(\bx)$ are simple to derive and are given below
\vspace{-4.0mm}
\begin{align}
\frac{\partial h_T}{\partial x_f} &= \begin{cases}
w_{\leaf} \sgn(\bx_{f_{v^*}} - t_{v^*}) & 
\begin{array}{c}
\hat{I}_{\leaf}(\bx) > 0 \\ 
f_{v^*} = f
\end{array} \\
0 & \text{otherwise}
\end{cases} \label{eq:treeDecisionDeriv} \\
\frac{\partial h_T}{\partial w_{\leaf}} &= \hat{I}_{\leaf}(\bx) \label{eq:treeWeightDeriv}
\end{align} \\[-5.0mm]
However, since only two gradient terms per tree are non-zero, it is inefficient to explicitly calculate the derivatives for all components. Instead the gradient can be efficiently calculated from the results of algorithm~\ref{alg:computeKeyAndSignedMarginTree} where $w_{\leaf} \sgn(r^*)$ gives~\eqref{eq:treeDecisionDeriv} for gradient component $f_{v^*}$ and $|r^*|$ gives~\eqref{eq:treeWeightDeriv} for gradient component $\leaf$. All remaining components thus have partial derivatives of $0$.

Random hinge ferns have a similar formulation as random hinge trees, except that where a depth $D$ hinge tree has $2^D - 1$ decisions, a random hinge fern has only $D$ decisions. Random hinge ferns and trees both have $2^D$ leaves. The traversal algorithm used for evaluating a prediction or gradient is similar, except the output vertex index $v^* \in \{ 0, 1, \hdots, D-1 \}$. %$ and given in algorithm~\ref{alg:computeKeyAndSignedMarginFern}.

\section{Random Hinge Forest and Optimization}
So far, discussions have covered individual trees and removed from learning tasks. A random hinge forest merely amounts to several random hinge trees and, unlike random forest and similar, the proposed machinery outputs the predictions of each individual tree. The machine learning practitioner decides how the predictions should be aggregated. Whether this is in the form of a mean, linear, semi-na{\"i}ve Bayes, etc... aggregation is left up to the user thus making random hinge forest a generalized learning component of computation graphs. That said, random hinge forest does require some careful initialization and suitable optimizers.

Each random hinge tree of a random hinge forest is initialized with random input feature indices $f_v$ and thresholds $t_v$. This is similar to using random subsets of features as was done in~\cite{kontschieder2015deep} for some experiments. Since the inputs are assumed to be the output of some learnable input transformation and can evolve into any form during the optimization, the feature selection performed in random forests seem less meaningful and we assume that it is not needed in end-to-end learning of random hinge forest. This assumption can actually be problematic and is discussed in section~\ref{sec:discussion}. As such, it is still important to carefully pick decision thresholds since, for example, should thresholds be too far from the input features, the min-max margin predictions produced by trees could saturate a softmax sigmoid used in softmax loss calculation, resulting in small gradients that could result in \textit{sluggish} learning. Likewise, in the spirit of random forests, we do desire a degree of variability between random hinge trees. To allow for arbitrary random initializations of random hinge trees, we use a modified form of batch normalization~\cite{ioffe2015batch} to normalize inputs of random hinge forest. Thus, thresholds can safely be randomly initialized as $t \sim \mathcal{U}(-3,3)$ where thresholds now represent standard deviations from the means of inputs. Our batch normalization keeps a running mean and standard deviation estimate that yields identical computations in forward passes during both training and testing. Consequently, the backward pass treats mean and standard deviation values used in the normalization as constants when calculating gradients.
%\vspace{-3.0mm}
%\begin{align}
%\hat{B}_i &= \frac{B_i - \mu_i}{\sigma_i} \\
%\frac{\partial \hat{B}_i}{\partial x} &= \frac{1}{\sigma_i} \\
%\mu_{i+1} &= \alpha_i \mu_i + (1- \alpha_i) \text{E}[B_i]  \\
%\sigma_{i+1} &= \alpha_i \sigma_i + (1 - \alpha_i) \sqrt{\text{E}[(B_i - E[B_i])^2]} \\
%\alpha_i &= \min \left \{ \alpha_{\max}, \frac{i}{i+1} \right \}
%\end{align} \\[-4.0mm]
%where $B_i,\ \hat{B}_i$ is the mini batch and normalized mini batch at iteration $i$ respectively and $\alpha_{\max}$ prevents the average and variance estimates from becoming \textit{stuck} in later iterations. We use $\alpha_{\max} = 0.99$ in this work. 
Likewise, the task-specific weights $w_{\leaf}$ can be real numbers or even arbitray tensors of real numbers. In the context of classification with a softmax classifier, it makes sense to initialize $w_{\leaf} = 0$ which would imply a random guess probability prediction by the softmax classifier. However, we found it is best to initialize $w_{\leaf}$ to be \textit{small} random numbers. We use $w_{\leaf} \sim \mathcal{N}(\mu = 0,\sigma = 0.01)$ in this work. This random hinge forest initialization is specified in algorithm~\ref{alg:initForest}.

\begin{algorithm}[bt!]
\caption{InitializeHingeForest}
\label{alg:initForest}

\begin{algorithmic}
\STATE {\bfseries Input:} number of trees $M$, tree depth $D$, number of input features $F$
\STATE {\bfseries Output:} decision thresholds $\bt \in \reals^{M \times (2^D-1)}$, feature indices $\bF \in \{ 0, 1, \hdots, F-1 \}^{M \times (2^D-1)}$, leaf weights $\bw \in \reals^{M \times 2^D}$
\FOR{$m=0$ {\bfseries to} $M-1$}
\FOR{$i=0$ {\bfseries to} $2^D-2$}
  \STATE Sample $\hat{f} \in \{0, 1, \hdots, F-1 \},\ \hat{w} \sim \mathcal{N}(\mu = 0, \sigma = 0.01),\ \hat{t} \sim \mathcal{U}(-3,3)$
  \STATE Set $\bF_{mi} \leftarrow \hat{f},\ \bw_{mi} \leftarrow \hat{w},\ \bt_{mi} \leftarrow \hat{t}$
  %\STATE Set $\bw_{mi} \leftarrow \hat{w}$
  %\STATE Set $\bt_{mi} \leftarrow \hat{t}$
\ENDFOR
  \STATE Sample $\hat{w} \sim \mathcal{N}(\mu = 0, \sigma = 0.01)$
  \STATE Set $i \leftarrow 2^D - 1,\ \bw_{mi} \leftarrow \hat{w}$
\ENDFOR
\end{algorithmic}

\end{algorithm}

\begin{figure}[bt!]
\centering
\includegraphics[width=2in]{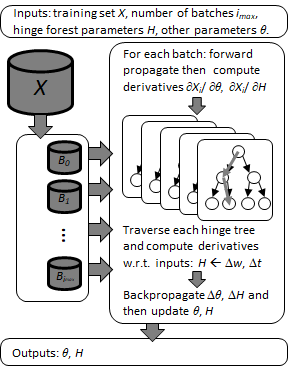}
\caption{Illustration of random hinge forest backpropagation from algorithm~\ref{alg:trainForest}.}
\label{fig:algorithm2}
\end{figure}

Once random hinge forest is initialized, then a gradient-like method can be used to minimize a task-specific loss function $\loss(x,y)$. Namely, the objective is to optimize the loss for the forest parameters $H^* = \{ (\bw_m^*,\bt_m^*) \}_{m=1}^M$ as well as any other parameters $\theta$ of a computation graph $g(\bx;H,\theta)$
\vspace{-2.0mm}
\begin{equation}
(H^*,\theta^*) = \argmin_{H,\theta} \left \{ \frac{1}{N} \sum_{n=1}^N \loss(g(\bx_n;H,\theta), y_n) \right \} \label{eq:forestGraphLoss}
\vspace{-2.0mm}
\end{equation}
where $N$ is the number of examples in the training set, and $M$ is the number of trees where each tree has leaf weights $\bw_m^*$ and decision thresholds $\bt_m^*$. The feature indices $\bF_m$ are fixed and not part of the optimization. We use stochastic gradient descent to optimize~\eqref{eq:forestGraphLoss} and backpropagation to calculate the gradient. A generic training algorithm for random hinge forest is illustrated in Figure~\ref{fig:algorithm2} with details given in algorithm~\ref{alg:trainForest}.

\begin{algorithm}[bt!]
\caption{TrainGraphContainingHingeForest}
\label{alg:trainForest}

\begin{algorithmic}
\STATE {\bfseries Input:} training set $X=\{\bx_n,y_n\}_{n=1}^N$, computation graph $g(\bx;H,\theta)$, batch size $N_B$, learning rate $\eta$, max number of gradient steps $i_{\max}$, initial depth $D$ random hinge forest parameters $H = \{ (\bF_m, \bw_m, \bt_m) \}_{m=1}^M$, initial other parameters $\theta$ \\
\COMMENT{\textbf{Disclaimer}: This is not a realistic procedure for stochastic gradient descent with backpropagation. The derivative calculation is decoupled from the other parameters to demonstrate how random hinge forest learns.}
\STATE {\bfseries Output:} learned parameters $H,\theta$
\FOR{$i=0$ {\bfseries to} $i_{\max}-1$}
  \STATE Draw mini batch $B \subseteq X$
  \STATE Forward propagate \\
    $\quad L = \frac{1}{N_B} \sum_{(\bx,y) \in B} \loss(g(\bx;H,\theta),y)$ 
  \STATE Compute derivatives w.r.t. to inputs for $\theta$ \\
  \COMMENT{Compute derivatives w.r.t. inputs for $H$}
  \FOR{$m=0$ {\bfseries to} $M-1$}
    \STATE Set $\Delta \bw_m = \bzero,\ \Delta \bt_m = \bzero$
    \FOR{$j=0$ {\bfseries to} $N_B-1$}
      \STATE Set $\bz = \text{input}_j$
      \STATE Compute \\
        $(\leaf,r^*,v^*)$ = HingeTreeTraversal($D, \bz, \bt_m, \bF_m$)
      \STATE Set $\Delta \text{input}_{jf_{mv^*}} \leftarrow \Delta \text{input}_{jf_{mv^*}} + w_{m\leaf} \sgn(r^*)$ %\COMMENT{From~\eqref{eq:treeDecisionDeriv}}
      \STATE Set $\Delta t_{mv^* j} \leftarrow - w_{m\leaf} \sgn(r^*)$ %\COMMENT{From~\eqref{eq:treeDecisionDeriv}}
      \STATE Set $\Delta w_{m\leaf j} \leftarrow |r^*|$ %\COMMENT{From~\eqref{eq:treeWeightDerive}}
    \ENDFOR
  \ENDFOR
  %\STATE Backpropagate $\Delta \theta = \nabla_{\theta} L,\ \Delta \bw = \nabla_{\bw} L,\ \Delta \bt = \nabla_{\bt} L$
  \STATE Backpropagate $(\Delta \theta, \Delta \bw, \Delta \bt) = \nabla L$
  \STATE Perform gradient step $\theta \leftarrow \theta - \eta \Delta \theta$
  \STATE Perform gradient step $\bw \leftarrow \bw - \eta \Delta \bw,\ \bt \leftarrow \bt - \eta \Delta \bt$
\ENDFOR
\end{algorithmic}

\end{algorithm}

The derivatives~\eqref{eq:treeDecisionDeriv}~\eqref{eq:treeWeightDeriv} produce extremely sparse gradients. Each training example will only result in two non-zero derivatives per tree; one for the decision threshold $t_{v^*}$ and one for the weight (or weight tensor) $w_{\leaf}$. The consequences of sparse gradients are not explored in this work. But such sparse gradients imply that the tree can be a \textit{sluggish} learner. And its backpropagated derivatives can make descendants of the computation graph also \textit{sluggish}. Therefore, it is important to use a suitable mini batch size (i.e. $> 1$) and advisable to use an adaptive learning rate gradient optimization algorithm. This work makes use of both AdaGrad~\cite{duchi2011adaptive} and Adam~\cite{kingma2014adam} for fast convergence. The gradient histories of tree thresholds, weights, and parameters of descendants of the forest will be comparatively small and both AdaGrad and Adam will provide larger learning rates for such parameters with sparse derivatives.

\begin{algorithm}[bt!]
%\caption{computeKeyAndSignedMarginTree}
\caption{HingeTreeTraversal}
\label{alg:computeKeyAndSignedMarginTree}

\begin{algorithmic}
\STATE {\bfseries Input:} tree depth $D$, feature vector $\bx \in \reals^F$, decision thresholds $\bt \in \reals^{2^D-1}$, feature indices $\bF \in \{ 0, 1, \hdots, F-1 \}^{2^D-1}$
\STATE {\bfseries Output:} leaf index $\leaf \in \{ 0, 1, \hdots, 2^D-1 \}$, signed margin $r^* \in \reals$, vertex index $v^* \in \{ 0, 1, \hdots, 2^D - 2 \}$
\STATE Initialize $\leaf = 0,\ v = 0,\ v^* = 0,\ r^* = \bx_{\bF_0} - \bt_0$
\FOR{$i=0$ {\bfseries to} $D-1$}
  \STATE Set $r = \bx_{\bF_v} - \bt_v$
  \IF{$|r| < |r^*|$}
    \STATE Set $r^* \leftarrow r,\ v^* \leftarrow v$
  \ENDIF
  \STATE Set $\leaf \leftarrow 2 \leaf + I(r > 0)$
  \STATE Set $v \leftarrow 2 v + I(r > 0) + 1$ 
\ENDFOR
\end{algorithmic}

\end{algorithm}

\section{Experiments}
\begin{figure*}[bt!]
\centering
\begin{subfigure}[t]{0.3 \textwidth}
\centering
\includegraphics[height=2in]{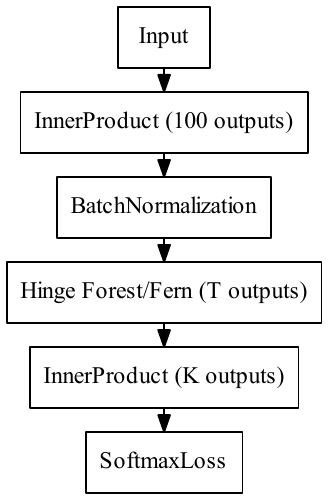}
\caption{Architecture used for UCI data.}
\label{subfig:Exp1Arch}
\end{subfigure}
\begin{subfigure}[t]{0.3 \textwidth}
\centering
\includegraphics[height=2in]{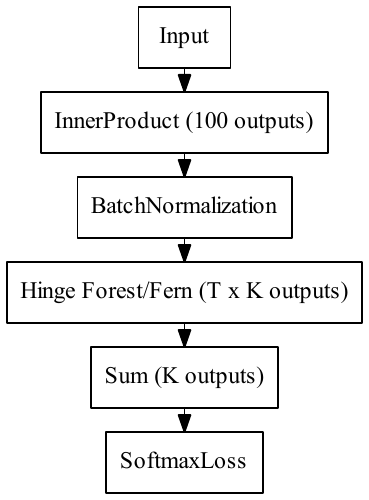}
\caption{Architecture used for letter and USPS.}
\label{subfig:Exp2Arch}
\end{subfigure}
\begin{subfigure}[t]{0.3 \textwidth}
\centering
\includegraphics[height=2in]{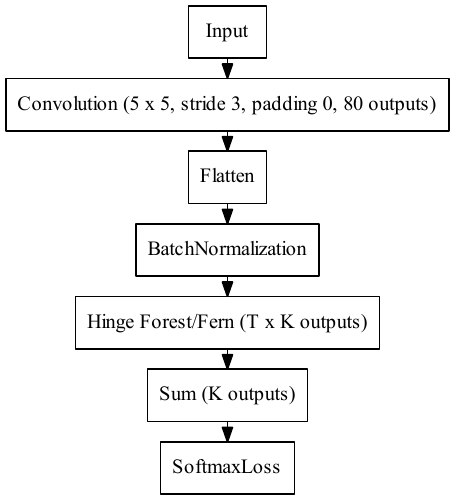}
\caption{Architecture used for MNIST.}
\label{subfig:Exp2ArchMNIST}
\end{subfigure}
\caption{Architectures used for results reported in this work. Here $T$ denotes the number of trees, $K$ denotes the number of class labels. The abalone regression experiment uses $\ell_2$ loss instead of the SoftmaxLoss.}
\label{fig:ExpArchs}
\end{figure*}

Two different sets of experiments were conducted to test and compare random hinge forest and ferns. The first set is run on several generic UCI~\cite{Lichman2013} data sets to show that both of the proposed solutions can generalize for a variety of learning tasks ranging from relatively small to large. We test configurations that vary the numbers of trees from $\{1, 10, 50, 100\}$ and the tree depths from $\{ 1, 3, 5, 7, 10 \}$. The best configuration is reported, and we compare against random forest~\cite{breiman2001random}, configured to $100$ depth $10$ trees. The data sets used in the UCI experiments are summarized in Table~\ref{tbl:Exp1Data} and Figure~\ref{subfig:Exp1Arch} depicts the network architecture used. All UCI experiments used AdaGrad~\cite{duchi2011adaptive} with a single dataset-specific learning rate over all configurations. 

The second set of experiments compares random hinge forest and ferns to~\cite{kontschieder2015deep,schulter2013alternating} on the letter~\cite{Lichman2013}, USPS~\cite{hull1994database}, and MNIST~\cite{lecun1998gradient} data sets. Following the experiments of~\cite{kontschieder2015deep,schulter2013alternating}, the tree number and depth were limited to $100$ and $10$ respectively. The data sets, experiment parameters and architectures used in this experiment are summarized in Table~\ref{tbl:Exp2} and Figures~\ref{subfig:Exp2Arch} and~\ref{subfig:Exp2ArchMNIST}. Experiments used the Adam~\cite{kingma2014adam} optimizer with $\beta_1 = 0.9$,  $\beta_2 = 0.999$, and $\eta = 0.005$. In contrast to the UCI experiments, these experiments report the best \textit{test error} over model states since~\cite{kontschieder2015deep} does not describe the use of validation sets. Lastly, to maximize our performance over MNIST, we report results for $1000$ trees/ferns. 

Unlike the work of~\cite{kontschieder2015deep}, random hinge forest does not have learnable representations built into decision nodes. Instead, random hinge forest relies on other layers to produce a pool of learnable feature representations. For both sets of experiments, except for MNIST, we use a pool of $100$ inner product features that are randomly assigned to random hinge forest decision nodes. For MNIST, we use a single layer of $80$ $5 \times 5$ convolutions with stride $3$ and no padding, which are then flattened before also being randomly assigned to decision nodes. 
\begin{table*}[bt!]
\caption{UCI data set properties. All data sets had given test sets except for iris. Training sets were randomly shuffled and a small portion was kept for validation (and also testing for iris).}
\vskip 0.15in 
\label{tbl:Exp1Data}
\centering
\begin{tabular}{|c|c|c|c|c|c|}
\hline
Data Set & Training & Validation & Testing & Dimension & Classes/Range \\
\hline
\hline
iris & 50 & 50 & 50 & 3 & 3 \\
abalone & 2089 & 1044 & 1044 & 8 & 3 \\
abalone regression & 2089 & 1044 & 1044 & 8 & $[1,29]$ \\
human activity & 5514 & 1838 & 2947 & 561 & 6 \\
madelon & 1500 & 500 & 600 & 500 & 2 \\
poker & 18757 & 6253 & 1000000 & 10 & 10 \\
\hline
\end{tabular}
\end{table*}

Both sets of experiments evaluated results over $10$ runs and the testing performances were averaged. The exception is iris, which does not possess a test set. Instead, iris was shuffled $5$ times and partitioned into $3$ folds, each used in turn as a training, validation, or test fold, totaling $15$ runs per configuration.

All experiments were conducted using an in-house and publicly available cross-platform C++14 framework (bleak)\footnote{Link, along with experiment scripts and details, shared upon publication.}. Random hinge forest should also be easily portable to other deep learning frameworks, such as Caffe. All experiments were conducted each using 1 CPU on a Windows 10 workstation running with an Intel Core i7-5930K (3.5 GHz) processor and 32 GB of RAM.

\section{Results}
The results of the UCI experiments are summarized in Table~\ref{tbl:Exp1Perf}. In the three experiments involving, abalone, abalone regression and human activity, the optimal random hinge forest and ferns configuration amounted to depth 1 hinge trees/ferns, or stumps. This makes reported errors for the two identical owing to the fact that depth 1 ferns and forests are identical learning machines. Importantly, for $5$ out of $6$ experiments, random hinge forest and ferns outperform random forest. On the madelon data set, however, random forest outperformed random hinge forest and ferns by a substantial $11.5\%$. This failure is discussed in Section~\ref{sec:discussion}.

The second set of experiments is summarized in Table~\ref{tbl:Exp2}. random hinge forest and ferns both outperform alternating decision forest and neural decision forest on letter, while random hinge forest is only slightly better than alternating decision forest on USPS but worse than neural decision forest. Random hinge forest and ferns are similar in performance to neural decision forest on MNIST although worse than alternating decision forest.

To test the limits of random hinge forest and ferns, we tested each using $1000$ depth $10$ trees/ferns on MNIST, reducing the error to $1.81\ (0.10)$ and $1.90\ (0.06)$, respectively. This greatly outperforms alternating decision forest and neural decision forest. Importantly, despite the large number of trees, it only took $\approx 3.5$ and $\approx 3$ hours to train $1000$-tree/fern random hinge forest and ferns, respectively, over $40000$ mini batches. All other parameters were kept the same. We also experimented with two nested convolutions and random hinge forest stacking, but performance did not exceed that reported for $1000$ trees/ferns. 

\begin{figure*}[bt!]
\centering
\begin{subfigure}[t]{0.51 \textwidth}
\centering
\includegraphics[height=2in]{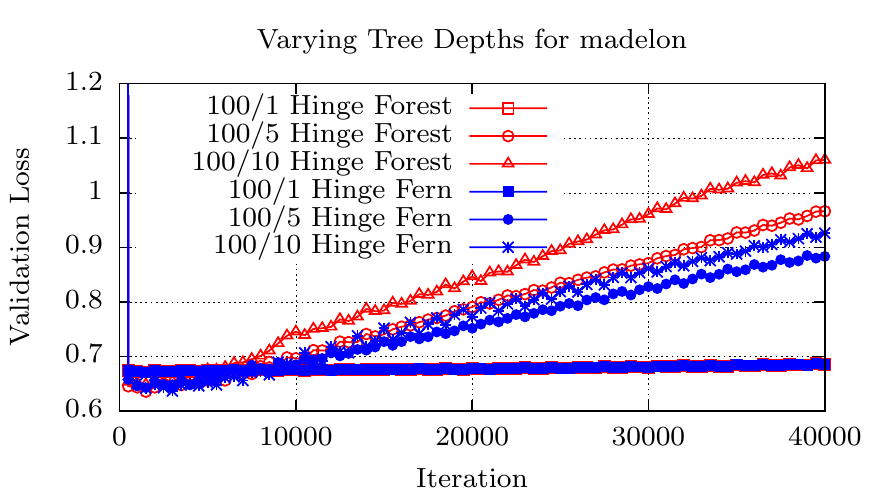}
\caption{Validation loss evolution for madelon.}
\label{subfig:varyingDepth}
\end{subfigure}
\begin{subfigure}[t]{0.48 \textwidth}
\centering
\includegraphics[height=2in]{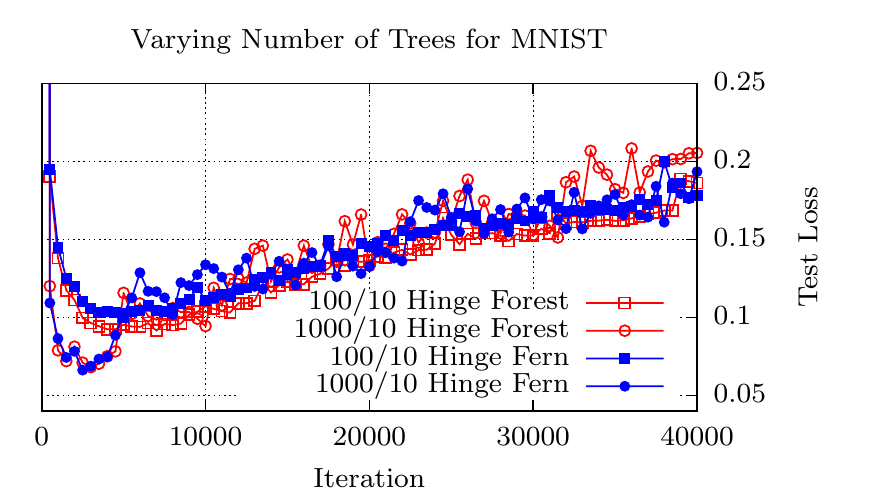}
\caption{Test loss evolution for MNIST.}
\label{subfig:varyingTrees}
\end{subfigure}
\caption{Validation/Test loss behaviors for varying tree depth~\ref{subfig:varyingDepth} and varying numbers of trees~\ref{subfig:varyingTrees}. Deeper trees and more trees both have the potential to generalize better, but can also severely overfit resulting in increasing validation/test losses over training iteration. Legends are to be interpreted as \# trees / \# depth.}
\label{fig:varyingParams}
\end{figure*}

Lastly, Figure~\ref{fig:varyingParams} illustrates the effects of varying tree depths and numbers of trees for both random hinge forest and ferns. These plots were each generated from the first run of the UCI madelon data set over tree depths $\{1, 5, 10 \}$ and the first run of the MNIST data set over the number of trees $\{ 100, 1000 \}$.

\begin{table*}[bt!]
\caption{Test errors or $R^2$ of the best performing models over number of trees $\{ 1, 10, 50, 100 \}$ and depths $\{ 1, 3, 5, 7, 10 \}$ compared to Random Forest on UCI data. The best configuration is expressed as $\text{\# trees}/\text{\# depth}$ and test error is reported as misclassification rate (standard deviation). The best performance for each data set is presented in bold.}
\vskip 0.15in 
\label{tbl:Exp1Perf}
\centering
\begin{tabular}{|c|cc|cc|c|}
\hline
Data Set & Hinge Forest Test & Config. & Hinge Fern Test & Config. & Random Forest \\
\hline 
\hline
iris & $\mathbf{2.13}\ (2.66)$ & 10/5 & $2.27\ (2.25)$ & 10/5 & $4.13\ (2.33)$ \\
abalone & $\mathbf{34.4}\ (0.56)$ & 50/1 & $\mathbf{34.4}\ (0.56)$ & 50/1 & $34.8\ (0.40)$ \\
abalone regression ($R^2$) & $\mathbf{0.57}\ (0.07)$ & 100/1 & $\mathbf{0.57}\ (0.07)$ & 100/1 & $0.54\ (0.01)$ \\
human activity & $\mathbf{5.52}\ (0.51)$ & 50/1 & $\mathbf{5.52}\ (0.51)$ & 50/1 & $8.23\ (0.50)$ \\
madelon & $40.8\ (1.60)$ & 10/3 & $40.8\ (1.41)$ & 50/7 & $\mathbf{29.3}\ (1.13)$ \\
poker & $\mathbf{32.6}\ (4.80)$ & 50/3 & $33.9\ (4.70)$ & 100/3 & $40.4\ (0.18)$ \\
\hline
\end{tabular}
\end{table*}

%
%iris & $2.13 \pm 2.66$ & 10/5 & $2.27 \pm 2.25$ & 10/5 & -- \\
%abalone & $34.4 \pm 5.62\e{-1}$ & 50/1 & $34.4 \pm 5.62\e{-1} $ & 50/1 & 45.5 \\
%abalone regression ($R^2$) & $5.66\e{-1} \pm 7.09\e{-2}$ & 100/1 & $5.66\e{-1} \pm 7.09\e{-2}$ & 100/1 & -- \\
%human activity & $5.52 \pm 5.07\e{-1}$ & 50/1 & $5.52 \pm 5.07\e{-1}$ & 50/1 & -- \\
%madelon & $40.8 \pm 1.60$ & 10/3 & $40.8 \pm 1.41$ & 50/7 & 46.1 \\
%poker & $32.6 \pm 4.80$ & 50/3 & $33.9 \pm 4.70$ & 100/3 & 43.2 \\
%

\begin{table*}[bt!]
\caption{Data set sizes, tree inputs, and classification error of random hinge forest, ferns, alternating decision forest~\cite{schulter2013alternating} and neural decision forest~\cite{kontschieder2015deep}. All features for the random hinge forest/ferns were simultaneously learned via backpropagation as part of random hinge forest/fern training. Test errors are expressed as average misclassification rate (standard deviation). The best performance for each data set is presented in bold.}
\vskip 0.15in 
\label{tbl:Exp2}
\centering
\begin{tabular}{|c|c|c|c|}
\hline
 & Letter~\cite{Lichman2013} & USPS~\cite{hull1994database} & MNIST~\cite{lecun1998gradient} \\
\hline
\hline
Training & 16000 & 7291 & 60000  \\
Testing & 4000 & 2007 & 10000 \\
Classes & 26 & 10 & 10 \\
Dimension & 16 & $16 \times 16$ & $28 \times 28$ \\
Features & 100 inner product & 100 inner product & 80 $5 \times 5$ convolutions \\
Batch Size & 53 & 29 & 53 \\
\hline
Random Hinge Forest & $\mathbf{2.56}\ (0.11)$ & $5.53\ (0.15)$ & $2.79\ (0.11)$ \\
Random Hinge Ferns & $2.78\ (0.12)$ & $5.64\ (0.24)$ & $2.85\ (0.13)$ \\
\hline
Alternating Decision Forest & $3.42\ (0.17)$ & $5.59\ (0.16)$ & $\mathbf{2.71}\ (0.10)$ \\
Neural Decision Forest & $2.92\ (0.17)$ & $\mathbf{5.01}\ (0.24)$ & $2.8\ (0.12)$ \\
\hline
\end{tabular}
\end{table*}

\section{Discussion}
\label{sec:discussion}
The UCI experiments demonstrate that random hinge forest and ferns can perform well on both relatively large and tiny data sets for a variety of learning tasks not usually attributed to differentiable learning. While not tabulated, the results over the varying numbers of trees and tree depths were often similar. In all but one case, both random hinge forest and ferns outperform random forest. The exception, \ie{}, the madelon~\cite{Lichman2013} dataset, is a synthetic binary classification problem generated from a number of Gaussian distributions as well as an abundance of useless noise features with a few other complexities added. We tried a number of different configurations, quantity of learnable features, weight decay, using Adam instead of AdaGrad, and feature selection, but we consistently obtained misclassification rates of $\approx 40.0$. In all tests, except for when using weight decay, the random hinge forest typically overfit the training data to perfection. Therefore, the most likely cause of poor performance is that inner product features are not suitable learnable features for the madelon data set. Where trees in random forest can deliberately choose split features as they grow, ignoring ones that do not prove useful to their task, hinge trees are assigned randomized learnable split features. Although these may be adjusted to fit its own randomized decision structure, they cannot be rejected. A hypothetical sparse learnable inner product feature might overcome this limitation as this would behave like the decisions used in trees of random forest. We leave this to future work.

The second set of experiments attempts to respect the constraints imposed on alternating decision forest and neural decision forest, where we limit ourselves to no more than $100$ depth $10$ trees. Importantly, while alternating decision forest are flexible enough to be employed with any loss function and neural decision forest can be incorporated within an arbitrary differentiable computation graph, random hinge forest and ferns are the first learnable decision tree machinery that can do both. In terms of performance, random hinge forest and ferns are comparable to alternating decision forest and neural decision forest. Our methods perform best on letter, second to neural decision forest on USPS, and second to alternating decision forest on MNSIT. Though, random hinge forest probably has the same performance as neural decision forest on MNIST. Noteworthy is that we maintain competitive performance on these three data sets with only a pool of $100$ learnable features for letter and USPS and $80$ learnable convolution kernels that produce $5120$ features when flattened on MNIST. By comparison, neural decision forest have a learnable innerproduct decision at every vertex of every decision tree, often resulting in many more parameters than random hinge forest. For instance, neural decision forest  uses $\approx 10 \times$ more parameters for USPS and about $\approx 17 \times$ more parameters for MNIST. random hinge forest use about $10 \%$ more parameters than neural decision forest for letter. Additionally, the competitive ferns have approximately half as many parameters as random hinge forest. 

Also, random hinge forest and ferns do not require such large mini batch sizes as used for neural decision forest, \eg{} 1000 for the MNIST dataset~\cite{kontschieder2015deep}. The large mini batch size for the latter is likely needed for the prediction node optimization when training, as the histograms have to be \textit{adequately} populated. However, random hinge forest can predict arbitrary weight vectors and are not limited to probability vectors and average leaf purity over the trees. Thus, random hinge forest can train with small batch sizes which, in combination with its logarithmic gradient calculation complexity, results in faster training and convergence. This efficient formulation, however, does limit random hinge forest to piecewise linear behavior while neural decision forest are naturally non-linear owing to its use of sigmoid decision functions and multiple leaves in prediction. These efficiencies translate into practical improvements. For example, if not for this efficiency, it would be prohibitively expensive to train $1000$ depth $10$ trees with learnable features on a single CPU. With OpenMP/C++11 threads and/or GPU acceleration, this could even be substantially faster.

\section{Conclusion}
Random hinge forest and ferns are efficient end-to-end learning machines that can be used to solve a variety of learning tasks on both relatively small and large data sets. Their ability to coexist in backpropagating computation graphs enables them to minimize a global loss function, be connected with any other differentiable component, and exploit learnable feature representations not afforded to the original random forest. This flexibility also comes with a robustness that can tolerate a variety of configurations and parameter initializations. We hope that random hinge forest can complement current deep learning machinery and maybe provide a simple means for new or experienced machine learning practitioners to develop and explore their own network architectures. 

\bibliography{paper}

\begin{thebibliography}{20}
\providecommand{\natexlab}[1]{#1}
\providecommand{\url}[1]{\texttt{#1}}
\expandafter\ifx\csname urlstyle\endcsname\relax
  \providecommand{\doi}[1]{doi: #1}\else
  \providecommand{\doi}{doi: \begingroup \urlstyle{rm}\Url}\fi

\bibitem[Amit \& Geman(1997)Amit and Geman]{amit1997shape}
Amit, Yali and Geman, Donald.
\newblock Shape quantization and recognition with randomized trees.
\newblock \emph{Neural computation}, 9\penalty0 (7):\penalty0 1545--1588, 1997.

\bibitem[Breiman(2001)]{breiman2001random}
Breiman, Leo.
\newblock Random forests.
\newblock \emph{Machine learning}, 45\penalty0 (1):\penalty0 5--32, 2001.

\bibitem[Criminisi et~al.(2010)Criminisi, Shotton, Robertson, and
  Konukoglu]{criminisi2010regression}
Criminisi, Antonio, Shotton, Jamie, Robertson, Duncan, and Konukoglu, Ender.
\newblock Regression forests for efficient anatomy detection and localization
  in ct studies.
\newblock In \emph{International MICCAI Workshop on Medical Computer Vision},
  pp.\  106--117. Springer, 2010.

\bibitem[Duchi et~al.(2011)Duchi, Hazan, and Singer]{duchi2011adaptive}
Duchi, John, Hazan, Elad, and Singer, Yoram.
\newblock Adaptive subgradient methods for online learning and stochastic
  optimization.
\newblock \emph{Journal of Machine Learning Research}, 12\penalty0
  (Jul):\penalty0 2121--2159, 2011.

\bibitem[Friedman(1991)]{friedman1991multivariate}
Friedman, Jerome~H.
\newblock Multivariate adaptive regression splines.
\newblock \emph{The annals of statistics}, pp.\  1--67, 1991.

\bibitem[Friedman(2001)]{friedman2001greedy}
Friedman, Jerome~H.
\newblock Greedy function approximation: a gradient boosting machine.
\newblock \emph{Annals of statistics}, pp.\  1189--1232, 2001.

\bibitem[Gall \& Lempitsky(2013)Gall and Lempitsky]{gall2013class}
Gall, Juergen and Lempitsky, Victor.
\newblock Class-specific hough forests for object detection.
\newblock In \emph{Decision forests for computer vision and medical image
  analysis}, pp.\  143--157. Springer, 2013.

\bibitem[He et~al.(2016)He, Zhang, Ren, and Sun]{he2016deep}
He, Kaiming, Zhang, Xiangyu, Ren, Shaoqing, and Sun, Jian.
\newblock Deep residual learning for image recognition.
\newblock In \emph{Proceedings of the IEEE conference on computer vision and
  pattern recognition}, pp.\  770--778, 2016.

\bibitem[Hull(1994)]{hull1994database}
Hull, Jonathan~J.
\newblock A database for handwritten text recognition research.
\newblock \emph{IEEE Transactions on pattern analysis and machine
  intelligence}, 16\penalty0 (5):\penalty0 550--554, 1994.

\bibitem[Ioffe \& Szegedy(2015)Ioffe and Szegedy]{ioffe2015batch}
Ioffe, Sergey and Szegedy, Christian.
\newblock Batch normalization: Accelerating deep network training by reducing
  internal covariate shift.
\newblock In \emph{International conference on machine learning}, pp.\
  448--456, 2015.

\bibitem[Kingma \& Ba(2014)Kingma and Ba]{kingma2014adam}
Kingma, Diederik~P and Ba, Jimmy.
\newblock Adam: A method for stochastic optimization.
\newblock \emph{arXiv preprint arXiv:1412.6980}, 2014.

\bibitem[Kontschieder et~al.(2015)Kontschieder, Fiterau, Criminisi, and
  Bulo]{kontschieder2015deep}
Kontschieder, Peter, Fiterau, Madalina, Criminisi, Antonio, and Bulo,
  Samuel~Rota.
\newblock Deep neural decision forests.
\newblock In \emph{Computer Vision (ICCV), 2015 IEEE International Conference
  on}, pp.\  1467--1475. IEEE, 2015.

\bibitem[Krizhevsky et~al.(2012)Krizhevsky, Sutskever, and
  Hinton]{krizhevsky2012imagenet}
Krizhevsky, Alex, Sutskever, Ilya, and Hinton, Geoffrey~E.
\newblock Imagenet classification with deep convolutional neural networks.
\newblock In \emph{Advances in neural information processing systems}, pp.\
  1097--1105, 2012.

\bibitem[LeCun et~al.(1998)LeCun, Bottou, Bengio, and
  Haffner]{lecun1998gradient}
LeCun, Yann, Bottou, L{\'e}on, Bengio, Yoshua, and Haffner, Patrick.
\newblock Gradient-based learning applied to document recognition.
\newblock \emph{Proceedings of the IEEE}, 86\penalty0 (11):\penalty0
  2278--2324, 1998.

\bibitem[Lichman(2013)]{Lichman2013}
Lichman, M.
\newblock {UCI} machine learning repository, 2013.
\newblock URL \url{http://archive.ics.uci.edu/ml}.

\bibitem[Ozuysal et~al.(2010)Ozuysal, Calonder, Lepetit, and
  Fua]{ozuysal2010fast}
Ozuysal, Mustafa, Calonder, Michael, Lepetit, Vincent, and Fua, Pascal.
\newblock Fast keypoint recognition using random ferns.
\newblock \emph{IEEE transactions on pattern analysis and machine
  intelligence}, 32\penalty0 (3):\penalty0 448--461, 2010.

\bibitem[Ronneberger et~al.(2015)Ronneberger, Fischer, and
  Brox]{ronneberger2015u}
Ronneberger, Olaf, Fischer, Philipp, and Brox, Thomas.
\newblock U-net: Convolutional networks for biomedical image segmentation.
\newblock In \emph{International Conference on Medical image computing and
  computer-assisted intervention}, pp.\  234--241. Springer, 2015.

\bibitem[Schulter et~al.(2013)Schulter, Wohlhart, Leistner, Saffari, Roth, and
  Bischof]{schulter2013alternating}
Schulter, Samuel, Wohlhart, Paul, Leistner, Christian, Saffari, Amir, Roth,
  Peter~M, and Bischof, Horst.
\newblock Alternating decision forests.
\newblock In \emph{Computer Vision and Pattern Recognition (CVPR), 2013 IEEE
  Conference on}, pp.\  508--515. IEEE, 2013.

\bibitem[Simonyan \& Zisserman(2014)Simonyan and Zisserman]{simonyan2014very}
Simonyan, Karen and Zisserman, Andrew.
\newblock Very deep convolutional networks for large-scale image recognition.
\newblock \emph{arXiv preprint arXiv:1409.1556}, 2014.

\bibitem[Su{\'a}rez \& Lutsko(1999)Su{\'a}rez and Lutsko]{suarez1999globally}
Su{\'a}rez, Alberto and Lutsko, James~F.
\newblock Globally optimal fuzzy decision trees for classification and
  regression.
\newblock \emph{IEEE Transactions on Pattern Analysis and Machine
  Intelligence}, 21\penalty0 (12):\penalty0 1297--1311, 1999.

\end{thebibliography}
\bibliographystyle{icml2018}

%%%% Put nothing here %%%%%

\end{document}